\documentclass[10pt,conference,review]{IEEEtran}
\IEEEoverridecommandlockouts
\usepackage{xcolor} 
\usepackage{multirow}
\usepackage{booktabs} 
\usepackage{graphicx}
\usepackage[belowskip=-12pt]{caption}
\usepackage{subcaption}
\usepackage{rotating}
\usepackage{tabularx}
\usepackage{wrapfig}
\usepackage{listings}
\usepackage{color, colortbl}
\usepackage{balance} 
\usepackage{lscape}
\usepackage[hidelinks]{hyperref}
\usepackage{xspace}
\usepackage{framed}
\usepackage{syntax}
\usepackage{soul}
\usepackage{flushend}
\usepackage{mathtools}
\usepackage{url}
\usepackage{lipsum}
\usepackage{amsmath,amssymb,amsfonts}
\usepackage{algpseudocode} 

\usepackage[tikz]{bclogo}
\usepackage[numbers,sort&compress]{natbib} 

\usepackage{hhline}
\usepackage{diagbox}
\usepackage{array, makecell}
\usepackage{microtype}
\newcommand{\tool}{\textsc{FairSense}\xspace}
\usepackage{enumitem}

\graphicspath{{./figures/}}
\newcommand{\tabref}[1]{Table~\ref{#1}}
\newcommand{\figref}[1]{Figure~\ref{#1}}

\newcommand{\secref}[1]{Section~\ref{#1}}

\definecolor{shadecolor}{rgb}{.9,.9,.9}

\def\BibTeX{{\rm B\kern-.05em{\sc i\kern-.025em b}\kern-.08em
    T\kern-.1667em\lower.7ex\hbox{E}\kern-.125emX}}

\begin{document}

\title{FairSense: Long-Term Fairness Analysis of ML-Enabled Systems}

\author{\IEEEauthorblockN{Yining She}
\IEEEauthorblockA{
\textit{Carnegie Mellon University}\\
yiningsh@andrew.cmu.edu}
\and
\IEEEauthorblockN{Sumon Biswas}
\IEEEauthorblockA{
\textit{Case Western Reserve University}\\
sumon@case.edu}
\and
\IEEEauthorblockN{Christian Kästner}
\IEEEauthorblockA{
\textit{Carnegie Mellon University}\\
}
\and
\IEEEauthorblockN{Eunsuk Kang}
\IEEEauthorblockA{
\textit{Carnegie Mellon University}\\
eunsukk@andrew.cmu.edu}
}

\maketitle
\thispagestyle{plain}
\pagestyle{plain}

\begin{abstract}
Algorithmic fairness of machine learning (ML) models has raised significant concern in the recent years. Many testing, verification, and bias mitigation techniques have been proposed to identify and reduce fairness issues in ML models. The existing methods are \emph{model-centric} and designed to detect fairness issues under \emph{static settings}.
However, many ML-enabled systems operate in a dynamic environment where the predictive decisions made by the system \textit{impact} the environment, which in turn affects future decision-making. Such a self-reinforcing \textit{feedback loop} can cause fairness violations in the long term, even if the immediate outcomes are fair.
In this paper, we propose a simulation-based framework called \tool to detect and analyze long-term unfairness in ML-enabled systems. 
Given a fairness requirement, \tool performs \textit{Monte-Carlo simulation} to enumerate evolution traces for each system configuration. Then, \tool performs \textit{sensitivity analysis} on the space of possible configurations to understand the impact of  design options and environmental factors on the long-term fairness of the system.
We demonstrate \tool's potential utility through three real-world case studies: Loan lending, opioids risk scoring, and predictive policing.
\end{abstract}


\section{Introduction}
\label{sec:introduction}

Socio-technical systems are increasingly using machine learning (ML) models to automate high-stakes decisions such as loan lending, drug risk scoring, predictive policing, college admission, and vaccine allocation \cite{jatho2022system, liu2018delayed, ensign2018runaway, farahani2021adaptive}. As unfair decisions made by such systems can cause harm to users and our society, approaches to developing fair systems have gathered significant interest in recent years. For example, researchers have developed various methods for fairness measures and identification \cite{galhotra2017fairness, biswas20machine,biswas21fair}, verification and testing \cite{aggarwal2019black, udeshi2018automated, galhotra2017fairness, zhang2020white, biswas23fairify}, and bias mitigation \cite{chakraborty2021bias, chakraborty2020fairway, nguyen23fix}.

Most of the existing work on fairness is \emph{model-centric} under \emph{static} settings, that is, it evaluates and improves fairness of a given model at a particular point in time. 
However, even if a system appears to be fair initially, fairness issues may arise after it has been deployed for a period of time; we call these \emph{long-term fairness} issues. A long-term fairness issue arises as a result of a \emph{feedback loop} between a system and its \emph{environment}~\cite{Sterman00,meadows2008thinking}: Decisions made by an ML-enabled system induce certain changes or \emph{shifts} in the environment, which, in turn, can influence the ensuing system behavior. For example, when an ML-enabled lending application declines a request for a bank loan, this decision may reduce the credit score of an individual, which further damages their chance of a future loan approval. 
If left unattended over a long period of time, such a \emph{self-reinforcing} feedback loop can result in discrimination against certain groups of individuals~\cite{o2017weapons}.

\textit{Unintended consequences} from feedback loops are being recognized as an emerging problem in ML-enabled systems~\cite{amodei2016concrete, o2017weapons, liu2018delayed, ensign2018runaway,d2020fairness, henzinger2023runtime}. Identifying long-term fairness issues poses new challenges beyond static fairness analysis, as it requires looking beyond the boundary of an ML model and analyzing possible \textit{interactions between the system and its environment} \cite{martin2020extending,pagan23classification,biswas2023towards}. 
The behavior of a system is influenced not only by ML system design decisions (e.g., data collection, agent policies, hyperparameters, optimization metrics, and retraining criteria) but also depends strongly on the dynamics of the surrounding environment (e.g., how people react to and adapt to the system and its decisions). Depending on the combination of these decisions and possible environmental dynamics, the system can evolve in numerous ways, some of which may result in an undesirable feedback loop. Thus, explicit consideration of the environment and its interactions with the system~\cite{Jackson95,GunterGJZ00} is crucial for identifying long-term fairness issues.

\begin{figure*}[!t]
    \centering
    \includegraphics[width=\linewidth]{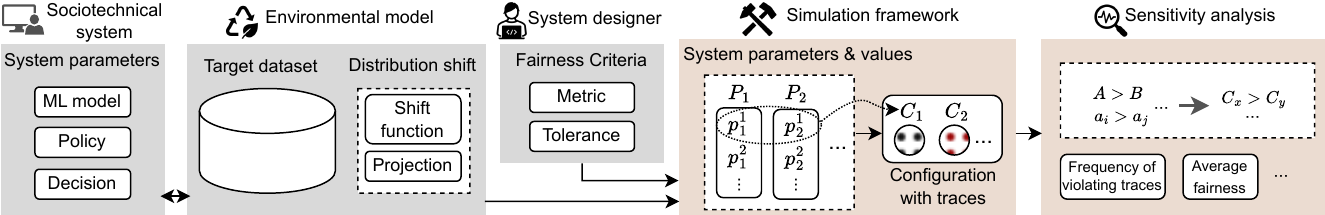}
    \caption{An overview of the \tool approach}
    \label{fig:overview}
\end{figure*}

In this paper, we propose \tool, a tool-assisted approach for \textbf{proactive analysis of long-term fairness issues that specifically considers a model of the environment and uncertainty about the environment}. A key distinction of our approach is that \tool is designed to aid developers early \textit{in the requirements and design stages}, so that they can focus on the design decisions and environmental factors that most impact long-term fairness, and avoid deliberating on others. That is, \tool helps to proactively analyze requirements and consequences of different designs before the system is implemented and deployed, which helps not only to create better initial designs, but also to plan for mitigations and monitoring to maintain fairness when the system is deployed in the real world.

We show an overview of the proposed framework in Figure~\ref{fig:overview}. A developer using \tool specifies three types of inputs: 
(1) \emph{system parameters}, which describe the space of configuration options (e.g., type of ML model and agent policies), to be explored, 
(2) desired \emph{fairness criteria}, and 
(3) an \emph{environmental model}, where the \emph{environmental parameters} control the dynamics of environmental changes that are induced by the system's decisions.
The latter model itself consists of (i) a \emph{target dataset} that represents the target population and (ii) a \emph{distribution-shift model}, which describes how the system outcome may cause changes in the dataset. The distribution-shift model is stochastic, explicitly encoding uncertainty about how the environment may evolve in response to system output. 

Given these inputs, \tool performs \textbf{Monte-Carlo simulation}~\cite{mooney97} to systematically generate \emph{traces} that show how the system and the environment may evolve together over time for a given \emph{configuration} (i.e., an assignment of values to the system and environmental parameters).
The desired fairness criteria are then evaluated over these traces, assigning each configuration a fairness metric that represents the level of unfairness that might arise over time. System designers are often overwhelmed with many design choices and can spend a lot of time negotiating a choice that ultimately matters little for fairness. \tool adopts \textbf{sensitivity analysis}~\cite{mcculloch2005sensitivity} to identify which system parameters and environmental parameters are the most influential to shape long-term fairness. This helps developers to focus their time effectively on reasoning about options that provide the highest leverage, e.g., monitor critical environmental parameters closely to reduce uncertainty and react in a timely fashion or invest in system design options that have the largest potential impact.
In addition, \tool enables a \textbf{trade-off analysis} between system utility and long-term fairness metrics, to aid developers in selecting decisions that achieve desired levels of utility and long-term fairness.

To demonstrate its potential utility, we have applied \tool on three real-world case studies, built on models and data from prior research: Loan lending \cite{liu2018delayed,d2020fairness}, opioids risk scoring \cite{10.1145/3442188.3445891, jatho2022system}, and predictive policing \cite{ensign2018runaway, akpinar2021effect}.
Our case studies show that \tool can be used to systematically analyze and understand the impact of design options on long-term fairness. 
The main contributions of the paper are:

\begin{itemize}
    \item A conceptual model of feedback loops and their impact on long-term fairness of ML-enabled systems (Section~\ref{sec:modeling}).
    
    \item A simulation-based framework that systematically explores possible evolution traces (Section~\ref{sec:simulation}) and performs sensitivity analysis to rank system parameters in terms of their impact on long-term fairness (Section~\ref{sec:sensitivity}).
    
    \item A prototype implementation of \tool and its demonstration on three real-world case studies (Section \ref{sec:evaluation-results}).
\end{itemize}

\section{Background}
\label{sec:background}

In this paper, we focused on the long-term fairness of ML-enabled sociotechnical systems. These \textit{systems} are software solutions that closely interact with humans and society in different domains, such as education, finance, and the judiciary.
The system consists of several components, such as the collected data, one or more ML models, and the decision-making entity. 
The system operates in a certain social context, which we refer to as the \textit{environment}.
The system and the environment interact continuously during its operation.
The environment dynamics can usually be decomposed further, such as the population distribution and human behaviors. The involvement of many \textit{agents}, such as system users and policymakers, affects various dynamics, which leads to uncertain evolution of the system and the environment over time.

A \textit{feedback loop} occurs when the system induces certain changes to the environment, which impacts the decision-making of the system through its input \cite{meadows2008thinking}. 
A \emph{balancing} feedback (or \emph{negative} feedback) loop is created by system structures that are sources of both stability and resistance.
On the other hand, a \emph{reinforcing} feedback (or \emph{positive} feedback) loop causes divergence and continuously shifts the environment toward a risky outcome,
commonly seen in various domains, such as biology, electronics, economics, and sociology \cite{kang2021role, henzinger2023runtime}.
An example of such a feedback loop is described in detail in \secref{sec:example} in the context of sociotechnical systems.
Identifying the cause of the feedback loop before deployment can help design interventions such as creating artificial feedback to mitigate existing one or choosing the best system parameter \cite{martin2020extending}. 
We proposed modeling the system, environment, and their interactions, which we call a \textit{feedback loop model}, to understand the system design space.

\begin{figure*}
    \centering
    \includegraphics[width=.84\linewidth]{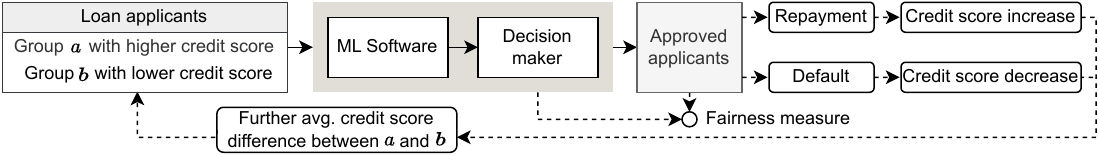}
    \caption{A feedback loop created by ML-enabled loan lending system}
    \label{fig:feedback1}
\end{figure*}

Various fairness criteria exist \cite{VermaR18}, with common metrics including including \emph{demographic parity} and \emph{equal opportunity} \cite{dwork2012fairness, hardt2016equality}.
For example, in a loan lending system, serving two groups $a$ and $b$, demographic parity measures the difference in the approval rates between individuals in $a$ and $b$. 
Formally, given the predictive outcome ($\hat{Y}$) and group membership feature ($A$), the demographic parity requirement is given by:
$\lvert P[\hat{Y} = 1 \vert A = a] - P[\hat{Y} = 1 | A = b] \rvert < \epsilon$ for some threshold  $\epsilon$.
In contrast, equal opportunity  measures the difference of true positive rates between the groups; formally, with $Y$ representing the actual outcome, 
$\lvert P[\hat{Y} = 1 \vert A = a, Y = 1] - P[\hat{Y} = 1 \vert A = b, Y = 1] \rvert < \epsilon$.

\section{Motivating Example}
\label{sec:example}

Unfairness in bank loan approvals and credit scoring has been investigated extensively in prior work \cite{Byrnesarchive2016, weber2020black,biswas23fairify}. 
For example, City National Bank was recently fined over \$31 million for discriminatory lending against Black and Latinos \cite{justice23}; and 
the Apple Card joint venture of Apple and Goldman Sachs has been accused of discriminating against female applicants for credit approval \cite{apple2019}. In the US, Equal Credit Opportunity Act of 1974 (ECOA) requires non-discrimination in lending.

A developer of a loan lending application can leverage existing fairness metrics or testing methods \cite{hardt2016equality, tramer2017fairtest, aggarwal2019black} to analyze whether the system \emph{statically} satisfies a fairness requirement at the model level. However, a seemingly fair loan lending system (e.g., satisfying demographic parity at the time of deployment) may begin to exhibit unfair behaviors over time. \figref{fig:feedback1} depicts a possible feedback-driven interaction between the ML-enabled system and the environment. The ML model here uses the applicants' credit score to predict the likelihood of on-time loan repayment; only if this predicted value is above a certain threshold, the applicant is granted the loan. 
Suppose that group $a$ historically has a higher average credit score than group $b$. To reduce this gap, a policy may deliberately approve a higher number of applications from group $b$. If, however, individuals in group $b$ are more inclined to a loan default, their average score may decrease at a higher rate than those in group $a$. Furthermore, an individual whose application gets rejected may begin to apply for other loans, incurring multiple \emph{hard inquiries} that further decrease their credit score~\cite{fico23}.
Thus, a feedback loop will influence the credit scores of members of $b$ to decrease over time. Even if the ML model satisfies demographic parity in the short-term, the system could begin to show unfair behavior over time due to the shifting distribution of credit scores.

The intensity of the feedback loop also depends on many factors, such as the magnitude of credit score decrease for a default and the loan approval threshold.
Some of these are configurable system parameters (e.g., the loan-approval threshold), and some are uncontrollable environmental parameters (e.g., credit score update model).
At design time, it may be challenging to understand how these different parameters might give rise to a feedback loop and negatively impact the long-term fairness of a system.
In the following sections, we describe our approach for explicitly modeling feedback interactions between the ML-based system and the environment, and a simulation-based analysis to understand the impact of both system and environmental parameters on long-term fairness.

\section{Modeling Feedback Loops}
\label{sec:modeling}

We propose a conceptual framework to specify the structure and key elements of feedback loops in an ML-enabled system. Then, we describe how this conceptual framework can be used to develop a design-time analysis for long-term unfairness.

\subsection{Feedback Loop Model} 

\begin{figure}[t] 
    \centering
    \includegraphics[width=\linewidth]{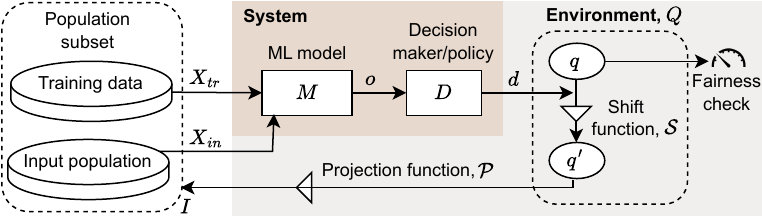}
    \caption{Feedback loop model of ML-enabled system}
    \label{fig:model}
\end{figure}

We illustrate the conceptual model of a feedback loop in an ML-enabled system in \figref{fig:model}: 
The system encompasses the ML model and any other components needed to produce the decisions; here, we modeled the system with the two components, the ML model and the \emph{decision maker}.
Feedback loops are system-level phenomena, and hence modeling the entire system, the environment, and their interactions is necessary to analyze these phenomena. 
mGiven input data (population sample $X_{in}$), the ML model $M$ generates predictions $o$. For example, in a loan lending system, $M$ takes the applicants' data as input, and outputs the probability of repayment. The decision maker entity $D$ (e.g., the bank) then makes decision $d$ (e.g., loan approval or rejection) based on the predictions $o$.

The environment is modeled as the stateful entity $Q$, where each state $q \in Q$ captures the relevant properties of the population at a certain point in time. 
We model $X_q$ as the population characteristic of the state $q$ such that each individual $x \in X_q$ has the attributes $\langle x_1, x_2, ... \rangle$.
For the loan lending application, $x_i$ can be the protected attribute (e.g., age, gender, race) or non-protected attribute (e.g., credit score, income, education) of the applicants. 

A system decision can affect the environment and \textit{change} certain attributes, such as credit scores. The change is modeled by a \emph{distribution-shift function} $\mathcal{S}: Q \times D \rightarrow \Delta(Q)$, where $\mathcal{S}$ is a stochastic function and $\Delta(Q)$ represents the probability distributions over the possible states $Q$. \emph{The stochastic nature of this function captures uncertainty about the way in which the environment evolves given a system decision.}
For some decision $d$, the environment shifts from $q$ to $q'$, where $q' \sim \mathcal{S}(q, d)$ represents the resulting probability distribution over the environmental states.
Over time, one possible evolution of states is $\langle q_0, q_1, q_2, ... \rangle$, which are impacted by the series of system decisions $\langle d_0, d_1, d_2, ... \rangle$ and where $q_n$ is sampled from the distribution $\mathcal{S}(q_{n-1}, d_{n-1})$.

The environmental state might not be fully observable by the system. In the feedback loop model, a \emph{projection function} $\mathcal{P}: Q \rightarrow I$ determines the observable parts, where $I$ serves as the input to the system. 
In practice, 
the developer may choose input data $X_{in}$ from $I$ in the next step. The training data $X_{tr}$ is optionally set aside from $I$ for updating the ML model periodically; 
i.e., $M$ can be left static or retrained over time.
In \tool, we model $\mathcal{P}$ as a stochastic sampling function,
since it can depend on various uncertain factors, such as human behavior, economic condition, and geographic location.
While the system is in deployment, the environment evolves through a series of distribution shifts and may reach a state that can cause $M$ to exhibit a certain level of unfairness.

\textbf{Example:}
In the loan lending example, the shift function ($S$) can be modeled using a stochastic function that changes the credit score of the individuals based on the decision, following a Normal distribution $\mathcal{N}(\mu,\,\sigma^{2})$. \tool employs separate distributions, $\mathcal{N}_1$ and $\mathcal{N}_2$ for approval and rejection decisions. In addition, \tool allows developers to conduct analysis for multiple environmental models by enumerating various options for the $\mu$ and $\sigma$, to reflect aggressive or conservative updates in the credit score. Similarly, $\mathcal{P}$ samples the input population by using a normal distribution $\mathcal{N}'(\mu,\,\sigma^{2})$.

\subsection{Feedback Loop Analysis}

Given an instantiation of the above feedback loop model for a particular system, \tool provides an analysis for understanding how (1) different ML system design options and (2) the dynamics of the environment may impact the long-term fairness of a system. The output of this analysis could be used by developers to (1) identify and select design options that improve long-term fairness (while considering trade-offs against other quality attributes, such as system utility) and (2) monitor the environment for the actual dynamics and apply interventions when necessary (e.g., modifying the decision-making policy)~\cite{farahani2021adaptive}.

Note that the analyst does not need to provide a perfectly accurate model of the environment for the analysis to be useful. The main objective is to identify what design decisions and environmental factors are important, not what exactly will happen in a particular environment. Where uncertainty exists (for instance, if it is unclear how strongly credit scores are impacted by declined loans), the analyst can model uncertain factors explicitly as \emph{parameters} to be explored by \tool{}.

For its analysis, \tool conducts a type of simulation-based \emph{configuration analysis}. Each component of a feedback loop model (i.e., $M, D, Q, \mathcal{S}, \mathcal{P}$) contains one or more of \emph{system} or \emph{environmental parameters}. System parameters (such as the choice of ML models and the approval threshold in loan lending) are decisions that are configurable by the developer, while environmental parameters (such as the credit score change mechanism for a loan default) are assumed to be uncontrollable but observable by the ML system. Each parameter is associated with a set of \emph{parameter values}; for example, the approval threshold for the loan lending policy may take on a value from a given range of parameter values (e.g., a credit score of 600).

Then, as an input to this analysis, the developer identifies a set of relevant system parameters (denoted $\mathbb{P}^s = P^s_1 \times P^s_2 \times \dots P^s_m$) and environmental parameters ($\mathbb{P}^e = P^e_1 \times P^e_2 \times \dots P^e_n$). The latter set of parameters and their values are typically elicited through a discussion with stakeholders (e.g. policy makers) or estimated based on prior data (e.g., a historical analysis of lending decisions and their impact on credit scores).
The combinations of different parameter values lead to a large number of \emph{configurations} ($C_1, C_2, ...$) for the feedback loop model, where each configuration $C_i = \langle p_{1}, p_{2}, \dots p_{m+n}  \rangle$ is a member of the space $\mathbb{C} = \mathbb{P}^s \times \mathbb{P}^e$. The idea behind \tool then is to simulate the feedback loop model under all possible configurations and extract relationships between the parameters and a desired long-term fairness measure. More precisely, we state the goal of the feedback loop analysis as: \emph{Given a feedback loop model of a system ($M, D, Q, \mathcal{S}, \mathcal{P}$) and its possible configurations ($\mathbb{C}$), which of the system and environmental parameters have the most impact on its long-term fairness, negatively or positively?}
\section{Simulation Framework}
\label{sec:simulation}

In this section, we describe how the feedback loop model is simulated for long-term fairness analysis.

\subsection{Monte-Carlo Simulation}

First, we obtain a target dataset that represents the environment. Taking the loan lending system as an example, the environment state can be represented by a snapshot of the dataset containing the credit scores of all loan applicants. The effect of the system decisions would be reflected in the changes in the dataset. 
For each configuration, we simulate the feedback loop model for $k$ time-steps and record one trace. In every time-step $t$, the system takes the inputs from the current state of the environment. The decisions of the system bring certain changes to the environment in the subsequent time-step ($t'=t+1$). Then, the system makes a new set of decisions based on the new inputs from the updated environment in step $t'$.
We record the inputs ($X_{in}$), outputs ($o$, $d$), and the environmental state ($q$) in each step, together constituting a snapshot $s$. Thus, after $k$ simulation steps, we generate a trace $T=\langle s_1, s_2, ... s_k \rangle$.

However, the feedback model may evolve in numerous ways for the same configuration, due to the uncertainty of the environmental parameters and the interactions between the system and the environment.
To systematically account for the uncertainty, we conduct a \emph{Monte-Carlo simulation} for each configuration. 
The Monte-Carlo simulation is a computational technique that uses random sampling to model complex systems and assess the impact of uncertainty. By generating a wide range of possible scenarios, it allows for the analysis of outcomes under varied conditions. Here, for each configuration $C_i$, we repeatedly conduct random simulation and collect a set of traces $\mathcal{T}_i=\{T_1, T_2, .., T_{m_i}\}$. The number of times the simulation needs to be conducted depends on the variance of the generated traces. 
The goal is to get a stable distribution of traces with respect to long-term fairness, and stop early for efficiency purposes. 
Specifically, we want to ensure that the estimated mean falls within the 5\% confidence interval of the true value with a probability of over 95\% \cite{gilman1968a}.
Therefore, the simulation for one configuration will be run repeatedly until the recorded set of traces $\mathcal{T}=\{T_1, T_2, .., T_{m}\}$ satisfies $\frac{Z_{0.95}Std(\mathcal{LF}(\mathcal{T}))}{Mean(\mathcal{LF}(\mathcal{T}))\sqrt{m}} < 0.05$, 
where $Z_{0.95}$ is the coefficient for 95\%-confidence level which is equal to 1.96 
and $\mathcal{LF}$ is the selected long-term fairness metric.

\begin{figure}[]
    \centering
    \includegraphics[width=.92\linewidth]{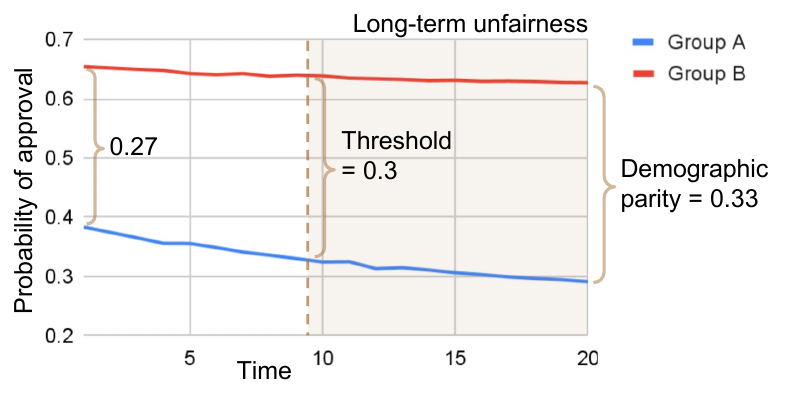}
    \caption{An evolution trace of loan lending system showing long-term unfairness.}
    \label{fig:motivating-example}
\end{figure}

\subsection{Long-term Fairness Evaluation}
Given a fairness criteria, we perform long-term fairness evaluation on all the traces. For example, one possible simulation trace for a configuration of the loan-lending system is shown in \figref{fig:motivating-example}; here, the feedback loop causes a divergence of fairness between two population groups and eventually results in a violation of the demographic parity requirements. 

We propose two different types of \emph{long-term fairness criteria} that can be used to evaluate a trace for the presence of  unfairness that arises over time:
\begin{itemize}[leftmargin=1em]
    \item \textbf{Average increase in unfairness: }This long-term fairness of a trace $T=\langle s_1, s_2, ... s_k \rangle$ is computed by measuring the average fairness of a trace using fairness criteria $\mathcal{F}$, e.g., demographic parity, and then subtracting the unfairness of the initial state. When the goal is to analyze the trend of long-term fairness (e.g., equilibrium, oscillation), this criterion would be a suitable choice:
    
    \vspace{-5mm}
    \begin{equation}\label{equ:avg increase}
        AvgInc_{\mathcal{F}}(T) = \frac{1}{k} \,\sum_{i =1}^{i = k} \mathcal{F}(s_i) - \mathcal{F}(s_1)
    \end{equation}

    \item \textbf{Maximum increase in unfairness:} This is measured by subtracting the unfairness of the initial state from the maximum unfairness exhibited by a trace for the given configuration. 
    When the goal is to avoid any severe unfairness in the future, analyzing this criterion would be useful:
    
    \vspace{-3mm}
    \begin{equation}\label{equ:max increase}
        MaxInc_{\mathcal{F}}(T) = \underset{x \in T}{\max} \mathcal{F}(x) - \mathcal{F}(s_1)
    \end{equation}
\end{itemize}
While we focus mainly on the above two criteria, the developer may plug in other criteria into \tool{}, such as \textit{bias promptness} (i.e., how quickly the system reaches a biased state) or \textit{violation frequency} (i.e., how many violations occur in a given time period).

\section{Sensitivity Analysis}
\label{sec:sensitivity}

Global sensitivity analysis technique has been successfully applied to investigate how uncertainty in the output of a model is attributed to different sources of model input \cite{saltelli2008global}.
This technique has also been applied to interpret the fairness in ML models \cite{grabowicz2022marrying, ghosh2023biased}. However, prior work focused on studying stationary ML models and evaluating the sensitivity of the training features on static fairness. Our approach, on the other hand, focuses on identifying parameters that are more likely to influence long-term fairness.
The sensitivity analysis on simulation results could be used by the developer to understand the long-term fairness impact of the entire \textit{design space} of system configurations. For example, if it is not clear how the environment behaves, different alternatives can be encoded as parameters, and a sensitivity analysis would identify whether that uncertainty in the environment is actually an important factor in influencing the long-term fairness of the model.

Exhaustively simulating the large space of configurations can become computationally prohibitive as the number of parameters and their possible values increases.
Hence, we propose a sampling heuristic to reduce the number of configurations to explore while preserving the accuracy required for effective sensitivity analysis.
In this section, we first describe the sensitivity analysis method and then the sampling heuristic.

\subsection{Sensitivity Analysis with Regression Modeling}
Regression analysis has been shown to be a good match for investigating parametric importance and sensitivity \cite{mcculloch2005sensitivity}. 
Compared with other sensitivity analysis methods like elementary effects methods and variance-based methods \cite{doi:https://doi.org/10.1002/9780470725184.ch3, doi:https://doi.org/10.1002/9780470725184.ch4}, the coefficient for each input variable in a regression model can be directly interpreted as the exact effect that the input variable brings to the output variable.

The idea is to learn a regression model that explains how the response is influenced by different options. In our case, the response is the long-term fairness measured in simulation and the options are possible values for the system and environment parameters. The model, a \textit{standardized multiple linear regression model}, is trained based on the measured fairness results of simulations with different configurations. The model's coefficients then indicate the influence of each parameter on the fairness result of the simulation.

The regression model has the structure shown below in Equation~\ref{math:linear regression}, where $y$ is the output variable, i.e., the long-term unfairness score. As introduced in previous section, $p_i$ is the option for the $i$th parameter in a configuration $C= \langle p_1, p_2, p_3, ..., p_n \rangle$. $\beta_i$ and $\beta_{i,j}$ are the coefficients for the terms. 
We include both individual terms, which assess the impact of each parameter ($P_i$) in isolation, and pair-wise interaction terms, which explore the combined effects of any two parameters (e.g., the interaction of $P_i$ and $P_j$). 
This dual approach also allows us to understand the independent contribution of every single parameter as well as the interplay between different parameters that influence long-term fairness. 
While interactions might be possible among more than two parameters at the same time in some case studies, our current focus is on the most salient pair-wise interactions, balancing the depth of analysis with model interpretability. 

\vspace{-3mm}
\begin{equation}\label{math:linear regression}
    y = \sum_{i = 1}^{n} \beta_i \cdot p_i + \sum_{i = 1}^{n}\sum_{j = i+1}^{n} \beta_{i,j} \cdot p_i \cdot p_j + \epsilon
\end{equation}

We use Analysis of Variance (ANOVA)~\cite{fisher1970statistical}  for the model, which allows us to identify which parameters contribute in a statistically significant way (using the common significance threshold of $p<0.05$). 
Then, we quantify the effect sizes of the statistically significant coefficients using the sums of squares and the eta-squared ($\eta^2$) derived from ANOVA. 
We evaluate whether an individual parameter (or its interaction with another parameter) is impactful based on Cohen's well-established guidelines \cite{cohen2013applied} ($\eta^2\geq0.01$, $\geq0.06$, and $\geq0.14$ indicate a small, medium, and large effect, respectively), and further rank them according to their effect sizes.
We additionally report the model's  $R^2$ value as a measure of fit, that indicates how much variance in the long-term fairness the model can explain in terms of the parameters and their interactions.

\subsection{Sampling Heuristic}
\label{subsec:sampling}

The presence of numerous parameters and their potential values  lead to an exponentially large configuration space. Given the time-intensive nature of Monte-Carlo simulations and the typical constraints of real-world development, simulation of all configurations can be challenging. To address this, we propose \textit{covering array sampling} on the configuration space to reduce the number of configurations to be explored, while ensuring a diverse and comprehensive coverage of the parameter values. 

Covering array sampling technique is widely used in software testing  \cite{Hartman2005} to achieve an adequate coverage of program behavior with a small number of carefully selected inputs. 
A covering array, characterized by a coverage number $g$, is a structured method to select combinations of a set of $n$ parameters' values. The key feature of a covering array is that it guarantees the inclusion of every possible combination of any set of $g$ factors' values from these $n$ parameters at least once; i.e.,  all possible $g$-factor interactions are covered within the array.
For example, in a 2-coverage array (i.e., $g=2$), an array of combinations is created such that every possible pair of parameter values is included.
This approach effectively minimizes the number of configurations \tool needs to simulate.
For example, a 2-coverage array for ten binary parameters can cover all pairwise interactions with only 12 configurations -- a significant reduction compared to the $2^{10}=1024$ configurations needed when enumerating all. 

\section{Experimental Setup}
\label{sec:experimental-setup}

To demonstrate the utility and applicability of \tool,
we conducted three case studies and answered the following research questions:
\begin{itemize}
\item{\textbf{RQ1.}} What are system and environmental parameters that significantly impact the long-term fairness of a system?
\item{\textbf{RQ2.}} What trade-offs among the parameters does \tool identify?
\item{\textbf{RQ3.}} How effective is the sampling heuristics in reducing the number of configurations explored while retaining the accuracy?
\end{itemize}

RQ1 is intended to demonstrate that \tool{} can potentially reduce the system developer's effort by identifying parameters that have significant influence on long-term fairness. RQ2 shows that \tool{} can be used by the developer to navigate the trade-off space between fairness and utility, to identify a design solution that acceptably meets both qualities. RQ3 evaluates the efficiency of \tool when the sampling heuristics is used.

\begin{table*}[t]
    \centering
    \caption{
    The parameters and their possible values. The middle and right column contain the configuration parameters taking part in the systems' decision-making and the interactions between systems and environments respectively.
    }
    \begin{tabular}{lll}
    \toprule
    \textbf{Case study}          & \textbf{System parameter}                                                                                             & \textbf{Environmental parameters}                                                                                                                                                                             \\\midrule
    \textit{Loan lending}       & \begin{tabular}[c]{@{}l@{}}Agent: max-util, eq-op\\ Bank utility func param: -10, -9 ... -3\end{tabular}               & \begin{tabular}[c]{@{}l@{}}Score update-repay: 8, 12 ... 20\\ Score update-default: -40, -32 ... -16\\ Shift function mode: expected, normal, aggressive\end{tabular}                             \\\addlinespace
    \textit{Opioid risk scoring} & \begin{tabular}[c]{@{}l@{}}Model type: XGBoost, MLP\\ Doctor threshold: 0.3, 0.4, 0.45, 0.5, 0.55, 0.6, 0.7\end{tabular} & \begin{tabular}[c]{@{}l@{}}Shift function mode - hospital visits: expected, equal, normal, aggressive\\ Shift function mode - prescription: expected, normal, aggressive\end{tabular}                      \\ \addlinespace
    \textit{Predictive policing} &  \begin{tabular}[c]{@{}l@{}}Model type: SEPP\\ Hotspot number: 50\end{tabular}                                                                                                                        & \begin{tabular}[c]{@{}l@{}}Discovery rate - hotspot cell: 0.8, 0.85 ... 1.0\\ Discovery rate - other cell: 0.2, 0.25 ... 0.5\\ Hot spot effect area range: 1, 2, 3\end{tabular} \\ \bottomrule
    \end{tabular}
    \label{tab:exp}%
    \vspace{-6mm}
\end{table*}%

The remainder of this section describes the case study systems and the experimental settings.\footnote{Further details of the case studies (i.e., ML model, decision maker, environmental dynamics, and configuration parameters) are presented in our supplemental material shared in the artifact: \url{https://github.com/cmu-soda/FairSense/tree/main/SupplementalMaterial}}
The space of possible configurations for each system is shown in \tabref{tab:exp}: In total, \textit{Loan lending} has 768 possible configurations, \textit{Opioid risk scoring} has 168, and \textit{Predictive policing} has 105. 
The code of \tool and the results of the case studies are available in our replication package.\footnote{\url{https://github.com/cmu-soda/FairSense/}}\looseness=-1

\subsection{Loan Lending} 

ML-enabled systems are used to predict the creditworthiness of people and approve or reject loan applications.
An ML model is trained on personal data (e.g., education, income, sex, race, credit score) of individuals, and then predicts the probability of an individual to repay or default. The decision-making system adopts a policy (e.g., a threshold) that approves or rejects loans based on the predictions. 
\textbf{Potential feedback loop:} 
\secref{sec:example} presented how a possible feedback loop might exhibit long-term unfairness against a group.
\textbf{ML-enabled system:} We leverage the predictive models defined by \cite{d2020fairness}, which determine credit score thresholds dynamically for different groups, and then employ the thresholds for loan approval and refusal.
\textbf{Environment:} We use the FICO dataset \cite{hardt2016equality} to present all the potential loan applicants. 
Detailed dynamics are described in \secref{sec:modeling}.A.

\textbf{Experimental settings:}
Following \citet{liu2018delayed} and \citet{d2020fairness}, we evaluated fairness between two racial groups -- White and Black. The fairness metric is given by the
demographic parity and the mean credit score difference between the groups. For long-term fairness metric, we leveraged the average increase (Eq. \eqref{equ:avg increase}) and the maximum increase (Eq. \eqref{equ:max increase}). 
The developer may choose to apply \tool using any of the defined long-term fairness metric; we used the maximum increase of demographic parity for the sensitivity analysis. To further explore the trade-off between long-term fairness and utility, we defined the profit of the bank as the utility metric. 

\subsection{Opioid Risk Scoring}

Opioids are a class of medicine that are frequently used for pain management. Common opioids such as oxycodone, morphine, fentanyl, and methadone, are used to reduce pain, but can cause overreliance and addiction, which is called Opioid Use Disorder (OUD).
Recent data shows a worsening situation with over 100,000 annual deaths from OUD for the first time in 2021 \cite{annual-death}. To reduce OUD, a Prescription Drug Monitoring Program (PDMP) is mandated in each state, which measures the opioid risk score of individuals. Many PDMPs use an ML-based application called NarxCare, which produces a numeric risk score (000-999) for individuals \cite{jatho2022system}. The risk score is shown to the doctors and pharmacists, based on which they can modify the prescription.

\textbf{Potential feedback loop:} The NarxCare ML model is trained using patients' medical records such as past opioid usage, number of pharmacies visited, number of prescribers, overlap from different prescribers, etc., \cite{narxcare-doc}. 
The NarxCare software uses these attributes to predict a risk score \cite{narxcare-doc}. Underrepresented patients including women and racial groups with complex medical conditions can have artificially inflated scores. Consequently, they can be denied a legitimate amount of opioids, suffer physical or mental debilitation, and be coerced into illegal activities \cite{jatho2022system}, which may further increase their risk score in the long term \cite{oliva2022dosing}.
\textbf{ML-enabled system:} 
The ML model adopted from \cite{vunikili2021predictive} uses patients' medical records to predict opioid risk scores and then doctors make prescriptions based on scores.
\textbf{Environment:} We use the publicly available Mimic-IV v2.2 dataset \cite{johnson2023mimic} to represent the potential patients. Patients will visit hospitals to get prescriptions; when a patient get an insufficient prescription, their further doctor visits may be affected. 
The shift function is defined to update the number of hospital visits by patients based on their  risk scores; higher the risk score a patient has, more visits they likely need~\cite{jatho2022system}.

\textbf{Experimental settings:}
We computed the fairness between two gender groups -- male and female. The long-term fairness metrics are the maximum increase (Eq. \eqref{equ:max increase}) in the gap between the mean opioid disorder risk scores of two groups, and the average increase (Eq. \eqref{equ:avg increase}) of the gap between ML model's performance metrics for two groups: Prediction accuracy and F1 score.
The first one is used for sensitivity analysis of the case study, and others are used in the trade-off analysis.
The utility metrics are defined by the daily average of ML model's performance metrics: prediction accuracy and F1 score.

\subsection{Predictive Policing}

Police departments are widely using ML algorithms to predict crime hotspots and deploy police based on the prediction. In a survey by the National Institute of Justice, over 70\% of agencies have reported to use predictive crime maps \cite{weisburd2008compstat}. 
\citeauthor{lum2016predict} experimented on the self-exciting point process (SEPP) model and data collected from Oakland, CA, to demonstrate over-policing in minority neighborhoods \cite{lum2016predict}.

\textbf{Potential feedback loop:} 
The model predicts crime hotspots based on the crime incident records from the past \cite{akpinar2021effect}.
Because of the crime increase for a short period or inaccurate prediction, more police may be sent to a certain location. This causes more crime discovery in that region, which, in turn, may influence the ML model to predict those regions as hotspots in the future.
\textbf{ML-enabled system:} The crime prediction model 
is adopted from \cite{akpinar2021effect}, which predicts crime intensity of each location for the next day. Based on the intensities, the decision maker allocates more police force to the top 50 cells (i.e., locations) with the highest intensities.
\textbf{Environment:} Adopted from \cite{akpinar2021effect}, we synthesized all the crime incidents that will take place. The shift function is used to derive the incidents occurring each day.
The projection function is defined as a stochastic function that determines which incidents would be discovered based on hotspot allocation; the incidents in the neighborhoods of hotspots are more likely to be discovered.

\textbf{Experimental settings:}
We measured the fairness of police allocation across districts by calculating the average pairwise Relative Percentage Difference (RPD) between districts' overpolicing scores. Following \citet{akpinar2021effect}, a district’s overpolicing score is determined by the relative number of predicted hotspots.
For the long-term fairness metrics, we computed the maximum and average increase of the district-wise allocation unfairness. 
The former metric is used for sensitivity analysis in our evaluation.
For system utility, we considered three metrics: The total number of discovered incidents, the mean of daily percentages of discovered incidents, and the number of correct predicted hotspots.


\section{Evaluation Results}
\label{sec:evaluation-results}

\subsection{RQ1: Sensitivity Analysis to Identify Impactful Parameters}
\label{subsec:sensitivity results}
We answer RQ1 through the sensitivity analysis results on the three case studies. We identify the most impactful parameters and their interactions, and show that only a small subset of the parameters are influential on long-term fairness.

\begin{table}[b]
    \setlength\tabcolsep{2pt}
    \centering
    \caption{
    Top 5 impactful regression terms for loan lending. A parameter can be numerical or categorical. Categorical parameters are transformed into binary variables using one-hot encoding. The specific values of the categorical parameter are listed in the second column.
     The pair of parameters represent interaction terms.}
    \begin{tabular}{llllll}\toprule
           & \textbf{Terms}      & \textbf{Dummies} & \textbf{Coefficient} & \textbf{Sum Sq.} & \textbf{$\eta^2$(\%)}\\ \midrule
    1 & Agent                           & max-util                 & -2.51E-02***    & 1.19E-01***  & 76.52       \\
    2 & Bank utility param                        &                  & ~8.45E-03***     & 1.35E-02***  & 8.72 \\
    3 & (Agent, bank utility p.)     & max-util                 & -8.38E-03***    & 1.35E-02***  & 8.69 \\
    4 & (Score update--d., agent) & max-util                 & ~3.35E-03***     & 2.16E-03***  & 1.39 \\
    5 & Score update--d.                    &                  & -3.38E-03***    & 2.16E-03***   & 1.39\\ 
    \bottomrule
    \end{tabular}
    {p-values: ***p $< .001$; **p $< .01$; *p $< .05$}
    \label{tab:loan-top5}
\end{table}

\paragraph{Loan Lending}
\tool collected 6,844 traces in total for 768 possible configurations. On average, every trace has 3.2\% increase in the unfairness (demographic parity) at the final time step compared to the initial step.
Around 45\% of the configurations have more than $5\%$ increase on average, and around 17\% of configurations have $\geq10\%$ increase on average, demonstrating long-term fairness issues in loan lending.

The fitted regression model explains the variance well ($R^2=0.970$). 
\tabref{tab:loan-top5} shows the terms (i.e., individual parameters and their interactions) with the top 5 effect sizes (sum of squares) in the regression model, all statistically significant.

\textbf{Regarding RQ1}, there are only 5 out of 15 terms that can be considered as impactful ($\eta^2\geq0.01$). Among them,
the choice of \textit{agent} is the most dominant factor ($\eta^2=0.76$) influencing long-term fairness; max-utility agent can greatly improve long-term fairness.
The \textit{bank utility parameter} and its interaction with the \textit{agent} together explain $\sim$17\% of the variance in the total sum of squares; both have a moderate influence on long-term fairness.
One standard deviation increase in \textit{bank utility parameter} can decrease long-term fairness by 0.00845, indicating that it is fairer if a bank makes loan decisions conservatively (i.e., smaller \textit{bank utility parameter}).
However, with the existence of max-utility agent, the individual effects of \textit{bank utility parameter}, \textit{score update parameters}, and \textit{shift function mode} will be offset. 
This illustrates how sensitivity analysis can highlight the small number of decisions (or sources of uncertainty of environment parameters) that require careful attention, where one factor dominates, four more have moderate influences, and 10 more are largely negligible.

\paragraph{Opioid Risk Scoring}
We collected 1,780 traces in total for 168 possible configurations. The average initial unfairness score (average risk gap between groups) for all configurations is 0.253\%. However, in the final step of the simulation, the average unfairness score increases to 2.859\%. More than 25\% of configurations have an increase of more than $3.2\%$.

\begin{table}[t]
    \setlength\tabcolsep{1pt}
    \centering
    \caption{Top 5 impactful terms for opioid risk scoring.}
    \begin{tabular}{p{1.5mm}p{21.5mm}llll}\toprule
        & \textbf{Terms} & \textbf{Dummies} & \textbf{Coefficient} & \textbf{Sum Sq.} & \textbf{$\eta^2$(\%)} \\ \midrule
    1 & ML model & XGBoost & ~1.12E-02*** & 6.35E-03*** & 96.80 \\ \addlinespace
    \multirow{2}{*}{2} & \multirow{2}{*}{\parbox{3cm}{(ML model, Shift \newline function--prscrptn.)}} & (XGBoost, normal) & ~2.53E-03*** & \multirow{2}{*}{6.28E-05***}  & \multirow{2}{*}{0.96}\\
                                &                                                                     & (XGBoost, aggressive)       & ~2.65E-03***     &                           \\ \addlinespace
    \multirow{3}{*}{3} & \multirow{3}{*}{\parbox{3cm}{Shift \newpage function--hospital}}                          & normal           & ~3.00E-04        & \multirow{3}{*}{3.92E-05***} & \multirow{3}{*}{0.60}\\
                                &                                                                     & aggressive       & ~5.70E-05        &                           \\
                                &                                                                     & equal            & -3.03E-04       &                           \\ \addlinespace
    \multirow{2}{*}{4} & \multirow{2}{*}{\parbox{3cm}{Shift function -- \newpage prescription}}                      & normal           & -2.37E-03***    & \multirow{2}{*}{3.82E-05***} & \multirow{2}{*}{0.59}\\
                                &                                                                     & aggressive       & -2.38E-03***    &                           \\ \addlinespace
    \multirow{3}{*}{5} & \multirow{3}{*}{\parbox{5cm}{(ML model, Shift \newline function--hospital)}}     & (XGBoost, normal)           & -4.15E-04*      & \multirow{3}{*}{1.92E-05***} & \multirow{3}{*}{0.29}\\
                                &                                                                     & (XGBoost, aggressive)       & -3.21E-04       &                           \\
                                &                                                                     & (XGBoost, equal)            & -1.77E-03***    &                           \\ 
                                \bottomrule
    \end{tabular}
    {p-values: ***p $< .001$; **p $< .01$; *p $< .05$}
    \label{tab:opioid-top5}
\end{table}

The fitted regression model again explains the variance very well ($R^2=0.995$). 
We show the ranking of the terms for this case study in \tabref{tab:opioid-top5}. 
\textbf{Regarding RQ1}, only 1 out of 10 parameters is impactful ($\eta^2\geq0.01$): the choice of \textit{ML model},
explaining around 97\% of all variance. Choosing XGBoost model would significantly amplify long-term unfairness.
Although the other terms have much smaller effects ($\eta^2<0.01$), we noticed that two environmental parameters in the configuration, \textit{Shift function--hospital} and \textit{Shift function--prescription} and their interaction terms with \textit{ML model} also have small but statistically significant influences, that developers might want to consider when tuning the model. 

\paragraph{Predictive Policing}
We explored 2,304 traces in total for 105 possible configurations. The average initial unfairness score is 0.739\%. In the last step of the simulation, the average score for all configurations increases to 21.522\%. Around 22\% of all configurations see an increase of more than $25\%$.

The fitted regression model again explains most of the variance ($R^2=0.738$, considered high in sensitivity analysis).
\tabref{tab:police-top5} summarizes the ranking of the terms' effect sizes for predictive policing. 
\textbf{Regarding RQ1},
2 of 5 terms ($\eta^2\geq0.01$) are considered impactful.
The incident discovery rate for areas except the hotspots (\textit{Discovery rate--other}) has the greatest impact on long-term fairness, explaining 69\% of the variance. The increase in \textit{Discovery rate--other} can greatly enhance long-term fairness.
The second impactful parameter is the incident discovery rate for hotspot areas (\textit{Discovery rate--hotspot}), explaining about 5\% of the variance. Compared to \textit{Discovery rate--other}, the increase in the discovery rate for hotspot areas has the opposite effect, it reduces long-term fairness.
This suggests long-term fairness might be improved if the police force is not only concentrated in hotspot areas predicted by the SEPP model but also partially distributed to other regions. 

\textit{Summary.} We observe that only a small subset (10-40\%) of the parameters impact ($\eta^2 > 0.01$) long-term fairness. These results demonstrate that \tool{} can be used to identify the most impactful parameters and allow the developer to allocate their design effort on them.

\subsection{RQ2: Trade-off Between Long-Term Fairness and Utility}
\label{subsec:trade-off}

The fairness requirement of the system can be defined using multiple metrics. In addition, the developer would typically care about the utility of the system, e.g., financial profit of the bank.
Optimizing exclusively for one long-term fairness metric may overlook this aspect, as the configuration yielding the lowest unfairness scores does not necessarily guarantee optimal utility. 
Therefore, it is essential to identify the trade-off between multiple fairness criteria and the utility of the system.
To this end, given the long-term fairness metric and utility metric, \tool finds a set of Pareto-optimal configurations.
These configurations represent scenarios where any improvement in utility would result in a decrease in long-term fairness, and vice versa. 
In this section, we demonstrate the trade-offs that exist in the case studies.

\begin{table}[t]
    \setlength\tabcolsep{1pt}
    \centering
    \caption{Top 5 impactful terms for predictive policing.}
    \begin{tabular}{rp{4.64cm}lll}\toprule
    \multicolumn{1}{l}{} & \textbf{Terms}                                          & \multicolumn{1}{l}{\textbf{Coefficient}} & \multicolumn{1}{l}{\textbf{Sum Sq.}}  & \multicolumn{1}{l}{\textbf{$\eta^2$(\%)}} \\ \midrule
    1                    & Discovery rate--other                                  & -3.46E-02*** & 1.26E-01*** & 68.92                            \\
    2                    & Discovery rate--hotspot                                            & ~8.99E-03***                         & 8.48E-03*** & 4.64\\
    3                    & (Discovery rt--hotspot, discovery rt--o.) &                  ~1.75E-03                            & 3.21E-04  & 0.18\\
    4                    & (Discovery rt--o., hotspot area range)    &                  ~9.40E-04                            & 9.30E-05   & 0.05\\
    5                    & Hotspot area range                                           & -7.18E-04                           & 5.40E-05   & 0.03\\ 
    \bottomrule
    \end{tabular}
    {p-values: ***p $< .001$; **p $< .01$; *p $< .05$}
    \label{tab:police-top5}
\end{table} 

\begin{figure*}[t]
    \centering
    \begin{subfigure}{0.3\textwidth}
        \includegraphics[width=\linewidth]{figures/loanlending/radar\_new.pdf}
        \caption{Loan lending case study.}
        \label{fig:loan-radar}
    \end{subfigure}
    \begin{subfigure}{0.3\textwidth}
        \includegraphics[width=\linewidth]{figures/opioid/radar\_new.pdf}
        \caption{Opioid risk prediction case study.}
        \label{fig:opioid-radar}
    \end{subfigure}
    \begin{subfigure}{0.3\textwidth}
        \includegraphics[width=\linewidth]{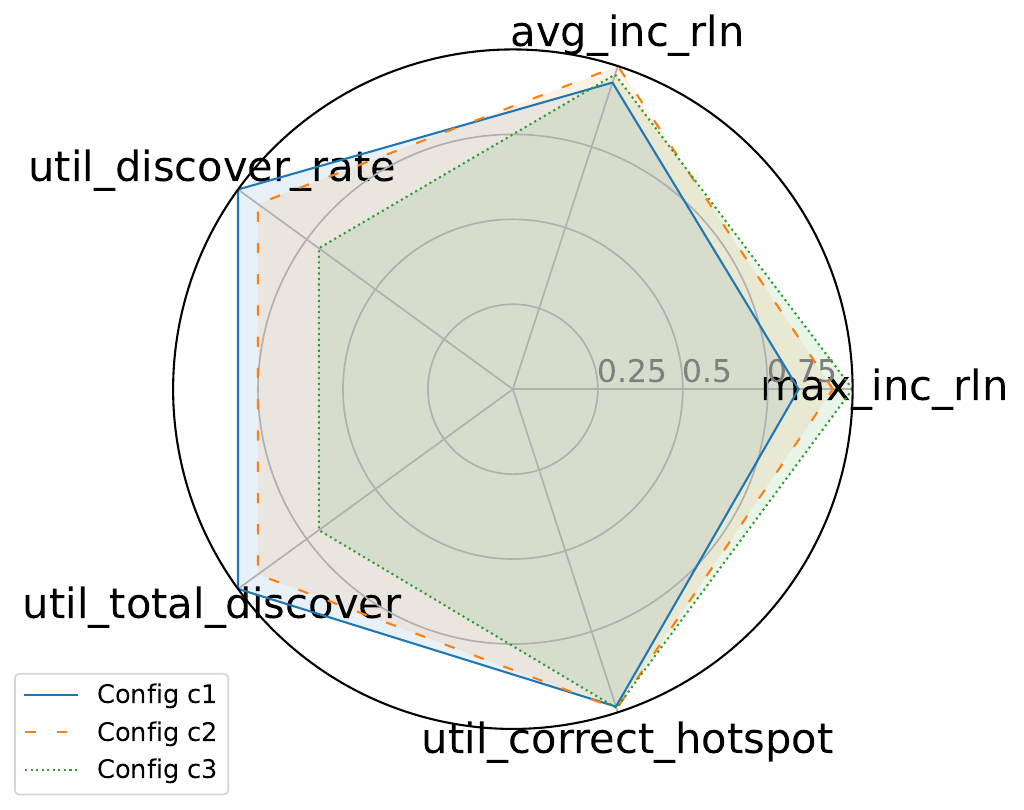}
        \caption{Predictive policing case study.}
        \label{fig:police-radar}
    \end{subfigure}
    \vspace{5pt}
    \caption{The radar plots visualizing trade-offs in three Pareto-optimal configurations for each case study. All values were scaled to [0,1]. A higher value implies better performance.}
    \label{fig:radars}
\end{figure*}

\paragraph{Loan Lending.}
\figref{fig:loan-radar} shows the trade-off among 4 long-term fairness metrics and 1 utility metric (defined in \secref{sec:experimental-setup}) for 3 Pareto-optimal configurations.\footnote{For visualization, we omitted showing all the Pareto-optimal configurations for the case studies in the plot. The assignments of all the configurations used in this section are presented in supplemental material.}
Configuration \textit{a1} optimizes maximum increase of demographic parity (\textit{max\_inc\_demo}) and financial profit of the bank (\textit{util\_profit}), while configuration \textit{a2} and \textit{a3} optimize average increase of credit score gap (\textit{avg\_inc\_credit}) and \textit{util\_profit}. 
A detailed examination of \textit{a2} and \textit{a3} reveals a discernible conflict between \textit{avg\_inc\_credit} and \textit{util\_profit}, highlighting the inherent trade-offs in optimizing these two metrics simultaneously. Interestingly, configuration \textit{a1} excels in both \textit{max\_inc\_demo} and \textit{util\_profit}, suggesting that there might be no evident conflict between them. 
If the developer places a higher emphasis on utility while optimizing all fairness metrics, \textit{a3} could be chosen, as it has nearly optimal utility and good fairness. 

\paragraph{Opioid Risk Scoring.}
\figref{fig:opioid-radar} shows the trade-offs for 3 Pareto-optimal configurations. Configuration \textit{b1} optimizes maximum increase in risk gap (\textit{max\_inc\_risk}) and utility of accuracy (\textit{util\_acc}). \textit{b2} is Pareto-optimal in both (i) optimizing average increase in accuracy gap (\textit{avg\_inc\_acc}) and \textit{util\_acc}, (ii) optimizing average increase in f1 gap (\textit{avg\_inc\_f1}) and \textit{util\_acc}. Notably, \textit{b3} optimizes two long-term fairness metrics \textit{max\_inc\_risk} and \textit{avg\_inc\_acc}.

Looking at \textit{b1} and \textit{b2}, we find that achieving the highest \textit{util\_acc} is possible while addressing any of the fairness metrics involved. 
Furthermore, \textit{b3} demonstrates that all three fairness metrics can attain favorable scores concurrently. 
Beyond these three Pareto-optimal configurations, it is surprising to find that there always exists a nearly optimal configuration for any pair of metrics. 
This observation suggests an absence of significant pairwise conflicts in this case study. 

\paragraph{Predictive Policing.}
\figref{fig:police-radar} illustrates the trade-offs for 3 Pareto-optimal configurations that optimize maximum increase of relative number of hotspots (\textit{max\_inc\_rln}) and total number of discovered incidents (\textit{util\_total\_discover}). 
The comparison of these 3 configurations reveals a conflict between the fairness metric (\textit{max\_inc\_rln}) and utility metrics (\textit{util\_total\_discover} and \textit{util\_discover\_rate}).
Notably, as we move from \textit{c1} to \textit{c2} and then to \textit{c3}, there is a monotonic increase in \textit{max\_inc\_rln}, accompanied by a corresponding decrease in both utility metrics. 
Furthermore, the magnitude of changes observed in these metrics appears to align with the principle of diminishing marginal utility. While \textit{c3} shows only a slight improvement in \textit{max\_inc\_rln} over \textit{c2}, it experiences a substantial reduction in the utility metrics. Conversely, \textit{c1} marginally outperforms \textit{c2} in utility, but at the cost of a significant reduction in \textit{max\_inc\_rln}.

\textit{Summary.} 
Overall, regarding RQ2, these observations highlight the complex and often delicate balance between maximizing long-term fairness and utility in system design, demonstrating the need for careful investigation of the trade-off between specific metrics during the design stage.

\begin{table*}[]
    \setlength\tabcolsep{5pt}
    \centering
    \caption{Comparison of efficiency of \tool with baseline method.}
    \begin{tabular}{llllllllllllll}\toprule

         {\textbf{Case Study}}                                      & \multicolumn{5}{c}{\textbf{2-coverage}}                                      & \multicolumn{5}{c}{\textbf{3-coverage}}                            & \multicolumn{3}{c}{\textbf{Baseline (no sampling)}} \\ \cmidrule(lr){2-6} \cmidrule(lr){7-11}\cmidrule(lr){12-14}
              & \textit{$R^2$}      & \textit{RBO}   & \textit{Tau}   & \textit{\# configs} & {\textit{time}}    & \textit{$R^2$}      & \textit{RBO}   & \textit{Tau}   & \textit{\# configs} & {\textit{time}}   & \textit{$R^2$}      & \textit{\# configs} & \textit{time}     \\\midrule
    {\textit{Loan lending}}       & 0.916 & 0.913 & 0.778 & 33        & {6min}    & 0.968 & 0.967 & 0.889 & 135       & {24min}  & 0.970 & 768       & 2.3hr    \\
    {\textit{Opioid risk scoring}} & 0.976 & 0.867 & 0.619 & 28        & {39min}   & 0.994 & 0.883  & 0.714 & 85        & {2hr}    & 0.995 & 168       & 3.9hr    \\
    {\textit{Predictive policing}} & 0.631 & 1.0  & 1.0 & 35        & {$\approx$39.5hr} & /        & /     & /     & 105       & {/}      & 0.738 & 105       & $\approx$118.6hr \\ \bottomrule
    \end{tabular}
    \label{tab:efficiency-compare}
    \vspace{-6mm}
\end{table*}

\subsection{RQ3: Performance Evaluation}
\label{subsec:efficiency}

In this section, we evaluate the efficiency of \tool in exploring a potentially large space of configurations for simulation through sampling.

Our hypothesis is that by applying covering array sampling, \tool can avoid simulating every configuration while maintaining a high accuracy of regression analysis and the identification of the most significant variables or interaction terms.
To test this, we consider the baseline as the ranking of the statistically significant (i.e., $p\le0.05$) terms, which is computed by analyzing every possible configuration from the given configuration space. 
Then, we compute the ranking of the same terms after applying both 2-coverage and 3-coverage sampling. We measure the time consumption of each sampling strategy, and computed their effectiveness using ranking similarity between the baseline ranks and the ranks found using sampling.
The ranking similarity metrics we used are  Rank Biased Overlap (RBO) with persistence 0.8 \cite{webber2010similarity} and Kendall Tau (Tau) \cite{665905b2-6123-3642-832e-05dbc1f48979}. 

Results for the three case studies, shown in \tabref{tab:efficiency-compare}, demonstrate that both 2-coverage and 3-coverage sampling require only a significantly smaller number of configuration simulations to obtain a regression model with little loss of model fit ($R^2$). The models trained on samples also identify impactful terms with a ranking very similar to the baseline derived from analyzing all configurations.
Specifically, the analysis results of 3-coverage sampling for loan lending and opioid risk scoring benchmarks are almost as good as the baseline while saving 80\% and 50\% simulation effort  (the predictive policing case study only has three parameters, so 3-way coverage samples all possible configurations). 
With 2-coverage sampling, simulation effort is reduced to 5\%, 16\%, 33\% compared to analyzing all configurations, while the resulting models still explain the variance similarly well.

In summary, to answer RQ3, the results show that the impactful parameters are still identifiable in the sensitivity analysis with carefully sampled configurations. The results in \tabref{tab:efficiency-compare} demonstrate 
that this heuristics can eliminate a significant number of configurations (and traces associated with them) while retaining the performance of fitted models.

\section{Threats to Validity}

\textbf{Fidelity of the environment models.}
The validity of simulation results depends on the fidelity of the environment model used for simulation. Instead of coming up with environmental parameters and dynamics on our own, we inferred these from the existing studies and analyses of these systems (loan lending: \cite{d2020fairness, liu2018delayed}; opioid risk scoring \cite{jatho2022system,vunikili2021predictive,10.1145/3442188.3445891}; predictive policing: \cite{ensign2018runaway,akpinar2021effect}). Although a process for developing environmental models is beyond the scope of this paper, we provide a discussion of existing methodologies that could be adopted for this purpose in Section~\ref{sec:discussion}.

\textbf{Validity of simulation results with respect to the real world.}
Simulation in \tool{} is a type of ``what-if'' analysis, estimating the potential impact of options for various systems and environmental parameters on fairness over time.

In this paper, we do not validate the accuracy of the simulation with regard to the real world.
Such validation would be very difficult and is orthogonal to the contributions of this paper. In theory, one way to validate the accuracy of our analysis results would be to deploy and execute a system under all possible configurations, and collect the resulting traces as the ground-truth data. Doing so, however, would be extremely challenging (and ethically questionable)~\cite{yin2024long}, especially for socio-technical systems where the system directly interacts with and influences users in the real world, sometimes negatively. To the best of our knowledge, no such ground-truth data is available for the kind of systems we study in this paper.

Instead, we provide a comparison against other prior analyses of long-term fairness for these systems. \emph{The comparison shows that the conclusions drawn from the simulation about the real world are consistent with those from the other analyses.} Although this does not offer the same level of validation of the simulator's ability to correctly model real-world behaviors against real-world observations, we believe that it provides evidence that the simulation is able to produce meaningful predictions about the potential impact of the parameters.

\emph{Loan lending:} Liu et al.~\cite{liu2018delayed} propose an analytical model that estimates the impact of different ML policies (e.g., equal opportunity vs. maximizing utility) on long-term fairness, studying a loan lending system based on the same dataset as the one used in this paper. 
Their model shows that the \emph{eq-op} agent, which optimizes for equality among different groups in the short-term, may actually result in long-term unfairness, in a somewhat counter-intuitive and surprising result~\cite[Theorem 3.4]{liu2018delayed}. Consistent with their conclusion, our sensitivity analysis results (Table II) also show that \emph{max-util} agent is the most effective in reducing long-term unfairness, and hence fairer than the \emph{eq-op} agent. 
Moreover, our analysis presents additional information, such as how the choice of agent together with other parameters, e.g., utility of the bank (Table-II, term 3), affect long-term fairness.

\emph{Predictive policing:} Ensign et al.~\cite{ensign2018runaway} propose an analytical model that estimates the occurrence of a feedback loop in the same predictive policing system that we studied. In particular, their model shows that over time, the system converges to a biased allocation scheme that assigns police only to the neighborhoods with the highest number of observed incidents and ignores those that are historically not hotspots. Their model also indicates that this bias can be mitigated by deliberately allocating resources to those non-hotspots, to increase the number of observations in those neighborhoods. These findings are consistent with our analysis results: The positive coefficient of discovery rate-hot spot (Table-IV, term 2) confirms that the dominant observation of incidents in hotspots would exacerbate unfairness~\cite[Sec. 4.2.1]{ensign2018runaway} On the other hand, the large negative coefficient of discovery rate-other (Table-IV, term 1) marks the importance of improving discovery rates of non-hot spot area for long-term fairness \cite[Sec. 4.2.2]{ensign2018runaway}.

\emph{Opioid scoring:} Although fairness in ML-based opioid risk scoring has been studied~\cite{10.1145/3442188.3445891, oliva2022dosing, McElfresh2023}, to the best of our knowledge, no prior work studies long-term fairness issues. 
\citet{adam2020hidden} conducted a simulation study on the same dataset that we used; specifically, they created artificial data drift and investigated the impact of different ML model retaining methods on performance. However, their work did not consider the impact of the system-decision on the environment and thus is not comparable to ours.

\section{Related Work}
\label{sec:related}

\textbf{Development of Fair ML-Enabled Systems.}
Understanding and improving algorithmic fairness in ML models has received significant attention in the recent past \cite{friedler2019comparative, biswas20machine, biswas21fair, tizpaz2022fairness, VermaR18, zhang2021ignorance, majumder2021fair}.
Many bias mitigation techniques have been proposed for ML algorithms \cite{chakraborty2020fairway,zemel2013learning, feldman2015certifying, zafar2015fairness, kamiran2012data, zhang2018mitigating, kamishima2012fairness, hardt2016equality, chakraborty2021bias}. The mitigation techniques can be categorized into preprocessing \cite{chakraborty2021bias, feldman2015certifying, kamiran2012data}, in-processing \cite{chen2022maat, zhang2018mitigating, kamishima2012fairness}, and post-processing methods \cite{hardt2016equality, grabowicz2022marrying}, depending on where the mitigation is applied.
However, several challenges remain for the development of fair ML-enabled systems \cite{biswas20machine,friedler2019comparative}.
Prior works showed that fairness-enhancing interventions can fail due to fluctuations in dataset characteristics, preprocessing methods, etc. \cite{qian2021my,biswas21fair, nguyen23fix,zhang2022adaptive}.
\citeauthor{holstein2019improving} outlined the challenges industry product teams face in developing fair systems \cite{holstein2019improving}. Thus, several software engineering techniques have been proposed for testing \cite{zheng2021neuronfair,udeshi2018automated,galhotra2017fairness,aggarwal2019black,tramer2017fairtest,fan2022explanation,10.1109/ICSE48619.2023.00136,soremekun2022astraea}, verifying \cite{zhang2020white,biswas23fairify,albarghouthi2017fairsquare,bastani2019probabilistic,john2020verifying,li2023certifying}, and achieving the accuracy-fairness trade-offs \cite{nguyen23fix, hort2021fairea}.
However, these works focus on fairness under static settings and do not consider long-term fairness.

\textbf{Long-term Fairness.} 
\citet{gohar2024long} conducted a survey on different notions of long-term fairness and created a taxonomy. 
\citet{d2020fairness} conducted simulations to show that static analysis is not sufficient to capture long-term fairness issues. Researchers focused on the predictive policing model to investigate the divergence of fairness over time \cite{ensign2018runaway,lum2016predict,akpinar2021effect}. 
Algorithmic solutions have also been proposed by considering the temporal factor of fairness in the sequential selection process \cite{hu2022achieving,hu2022achieving,wen2021algorithms,mouzannar2019fair}.
\citet{albarghouthi2019fairness} proposed a runtime specification language to monitor fairness statistics and provide warnings for violations. \citet{henzinger2023monitoring,henzinger2023runtime} built retrospective analysis and proposed a runtime statistical estimator to avoid long-term unfairness.
Several ML algorithms have been proposed for optimizing a long-term fairness objective under certain assumptions or fixed environmental dynamics \cite{weber2022enforcing, yin2024long, du2024long}.
However, no prior work has focused on analyzing the influence of system parameters on long-term fairness.
Understanding the dynamics of fairness requires modeling the system and its context \cite{schwobel2022long,selbst2019fairness,farahani2021adaptive}, and difficult to achieve through static analysis \cite{henzinger2023runtime}. 

\textbf{Feedback Loops.}
An emerging problem for ML systems is to ensure robustness in presence of feedback loops \cite{biswas2023towards, kang2022requirements, hellerstein2004feedback, amodei2016concrete,adam2020hidden}. \citet{o2017weapons} explained several harmful feedback loops in sociotechnical systems at length. 
\citet{pagan23classification} classified the different types of feedback loops in ML-enabled systems.
With an emphasis on accuracy, most of the ML research in the area focuses on data bias and distribution shifts induced from feedback \cite{taori2023data,krueger2020hidden,quinonero2022dataset}. 
However, designing an ML system for long-term fairness would need adaptive design and mitigation strategies \cite{casimiro2021self}. Recently, \citet{ReaderNPPVF22} proposed a system theory-based approach to quantify feedback in sociotechnical systems. 
\citet{martin2020extending} also recommend system-level analysis and in-depth understanding of the societal context to identify feedback loops. To that end, we have built a simulation-based framework to analyze long-term fairness issues.

\section{Discussions}
\label{sec:discussion}

The simulation-based analysis in \tool{} relies on an environmental model that describes how the environment evolves in response to the system output. \tool{} is specifically designed to enable reasoning about interactions between the system and the environment when certain details about the environment are unknown at the design time; these details can be encoded as  environmental parameters, which are then explored by the tool to provide insights into their impact. This allows the system designer to identify which uncertainty in the environment is most important to focus their efforts on, rather than wasting efforts on aspects of the environment that matter little for fairness. That is, the challenge of creating an accurate environment model is not a limiting factor, but a key motivation for sensitivity analysis in \tool{}.

Although our environmental models were derived from prior work (as described in Section~\ref{sec:experimental-setup}), in practice, creating environment models would involve a requirements engineering process understanding relevant environment behaviors from stakeholders and domain experts. Beyond requirements engineering techniques, the \emph{system dynamics} community has long studied a rich set of methods for building, simulating and analyzing environmental models in socio-technical systems~\cite{meadows2008thinking,Sterman00}. One promising method is a type of modeling notation called \emph{causal loop diagram (CLD)}, which is used to model the environment as a set of \emph{variables} (parameters in our terminology), relationships between those variables (i.e., shift dynamics), and possible feedback loops that arise from them~\cite{sterman01}.  CLDs have been used to model and simulate the environment in a wide range of domains, such as economics, social sciences, ecology, and public policy; methodologies for developing CLDs are also well-studied~\cite{DhirasasnaS19,sdr.1480,burns2001structural,sdr.280}.
CLDs have also been adopted in requirements engineering (to model the impact of software on sustainability, for example~\cite{PenzenstadlerDV18}), although not yet in the context of fairness, as far as we know.

A complementary approach to improving the quality of the models used in \tool{} is \emph{runtime monitoring}. Once a system is deployed, observations collected from the environment (i.e., new data samples) could be used to evaluate whether the environment model is consistent with the actual environmental behavior. If there are discrepancies (possibly due to inaccurate modeling or data drift), this information could be used to update the model and improve its fidelity. Recent work in runtime monitoring for fairness~\cite{henzinger2023runtime,henzinger2023monitoring,albarghouthi2019fairness} could be adopted for this purpose as part of a framework like \tool{}.

\section{Conclusions}
\label{sec:future}

Understanding the impact of design decisions for  ML-enabled systems has received much attention recently. Many ML interventions have been proposed to improve algorithmic fairness in static settings; however, the long-term impact of such interventions is still unclear, as the system interacts with the environment over time, possibly forming unexpected feedback loops. 
Precisely understanding the environment to accurately predict long-term fairness is challenging. To this end, we have proposed \tool, to aid developers in identifying and understanding design
decisions and environmental factors that impact long-term fairness.

\section*{Acknowledgments}
We'd like to thank our anonymous reviewers for their feedback, which greatly helped improve this paper. This work was supported in part by the National Science Foundation Awards No. 2131477, 2144860, and 2233871.

\renewcommand{\bibfont}{\footnotesize}
\bibliographystyle{IEEEtranN}
\bibliography{refs}

\end{document}